\documentclass{ws-procs9x6}
\newtheorem{example2}{Example}

\begin{document}
	
	\title{QUASI-DILEMMAS FOR ARTIFICIAL MORAL AGENTS}
	\author{D. KASENBERG$^*$, V. SARATHY, T. ARNOLD, and M. SCHEUTZ}
	
	\address{Human-Robot Interaction Laboratory, Tufts University,\\
		Medford, MA 02155, USA\\
		$^*$E-mail: dmk@cs.tufts.edu\\
		hrilab.tufts.edu}
	
	\author{T. WILLIAMS}
	\address{MIRROR Laboratory, Colorado School of Mines,\\
		Golden, CO 80401, USA\\
		E-mail: twilliams@mines.edu\\
                mirrorlab.mines.edu}
	
	\begin{abstract}
		In this paper we describe moral quasi-dilemmas (MQDs): situations similar to moral dilemmas, but in which an agent is unsure whether exploring the plan space or the world may reveal a course of action that satisfies all moral requirements.  We argue that artificial moral agents (AMAs) should be built to handle MQDs (in particular, by exploring the plan space rather than immediately accepting the inevitability of the moral dilemma), and that MQDs may be useful for evaluating AMA architectures.
	\end{abstract}
	
	
	\bodymatter
	
	\section{Introduction}\label{sec:intro}
        
	Much of the focus in developing and evaluating artificial
        moral agents (AMAs) has centered on
        moral dilemmas, here defined as situations in which the agent
        must choose one of a few courses of action, each of which
        nontrivially violates moral norms\cite{anderson2014geneth,kim2018computational,aaai18,muntean2014artificial,wallach2010conceptual}.  Nevertheless, some situations may \textit{appear} to the agent to be moral dilemmas (in that all of the `obvious' courses of action violate moral requirements), but may actually be solvable with a little ingenuity. Importantly, an agent will often not know \textit{a priori} whether a given moral problem has a solution --- they have not determined a solution to the problem, but cannot be certain that none exists.  We will refer to such problems as \textit{moral quasi-dilemmas} (MQDs).  The following is an example of a MQD:
        
        	\begin{example2}
        		{\it You are in the cockpit of a train approaching five railroad workers, who will die when the train hits them.  You have tried applying the brakes without success.  In front of you is a switch which you know can reroute the train onto a different track, but there is one person on this track as well.  It will take roughly ten seconds to reach the junction, after which you will be unable to reroute the train.  What do you do?}
        	\end{example2}
        	
        	Regardless of
        	whether a human placed into the above context 
        	ultimately chooses to eventually flip the switch, a rich environment
        	may be available for them to explore in the ten seconds before their
        	choice must be made. The cockpit of the train, for example, may
        	include many buttons and levers could affect the
        	situation in unknown ways. Some of these
        	potentially unexplored environmental features may afford some means of producing a
        	better outcome than the two presented by the switch, perhaps even
        	preventing \emph{all} deaths. Should the human operator spend some or all of
        	the remaining time searching for such a solution? How much time
        	and energy should they spend considering new courses of action, and to what extent should they attempt to physically explore their environment, before selecting the `lesser evil'?
	
        The remainder of the paper proceeds as follows. In
        Sec.~\ref{sec:rw} we discuss related work
        on moral dilemmas and AMAs. In
        Sec.~\ref{sec:mqd}, we then define and characterize our
        conception of MQDs. Next, in
        Sec.~\ref{sec:example}, we analyze a second example MQD. Finally, in
        Secs.~\ref{sec:why}-\ref{sec:conclusion} we discuss
        \emph{why} AMA designers should consider MQDs,
        \emph{how} agents able to handle MQDs might be
        designed, and conclude with possible directions for future work.
        
	\section{Related work}\label{sec:rw}
	
	Trolley problems and other moral dilemmas have been criticized as being unrealistic in various ways\cite{Kuipers2016}.  These critiques often misread the purpose of trolley problems, which is to use a purposely contrived scenario to elucidate key features of human moral judgment and decision-making.  Nevertheless, in practical situations, being too quick to treat a situation as a moral dilemma can cause one to miss out on creative ways to ``escape'' the dilemma.  Foot notes that in many moral dilemmas it is ``up to the agent to rack his brains for a way out before declaring that the conflict is real''\cite{Foot1983}.
	
	Outside of machine ethics, applied ethics (including engineering and business ethics) often emphasizes attempting to find solutions to apparent moral dilemmas.\cite{weston2006creative,Werhane1998,whitbeck2011ethics,kidder1995good}    Ethicists have framed this in terms of creative problem-solving\cite{weston2006creative}, transcending conceptual schemas\cite{Werhane1998}, applying design ideas to ethical problems\cite{whitbeck2011ethics}, and considering ``trilemma'' options\cite{kidder1995good}.

	Despite this, AMA architectures tend not to consider the possibility of ``escaping'' moral dilemmas.  Such architectures tend to operate either on problems explicitly assumed to be genuine moral dilemmas\cite{anderson2014geneth,kim2018computational}, 
	 or in simulated worlds sufficiently small and well-known that their solvability can be conclusively determined\cite{aaai18} 
	 .  Muntean and Howard describe creativity as being important to their AMA architecture, but it is not clear that their approach would be suitable for moral quasi-dilemmas\cite{muntean2014artificial}. Approaches based on cognitive architectures such as LIDA\cite{wallach2010conceptual} may hold some promise for this task (such architectures often already aim to model creativity), but so far this has not been the focus of these architectures.
	
	The present paper builds upon a blog post by Daniel Hicks, in which he describes the basic quasi-dilemma premise as a problem generally missed by conventional `principle-based' AMA architectures.\cite{hicks2014}  We consider this idea in greater detail, more precisely characterizing the problem, and examining its dimensions and its utility in designing and evaluating AMAs.

\section{Characterizing moral quasi-dilemmas}\label{sec:mqd}	
	
We define a \textit{moral quasi-dilemma} (hereafter \textit{MQD}) as a situation in which an
agent (1) is aware of multiple courses of action (one of which may
be inaction), each of which (or the outcomes thereof) violates some subset of the agent's
moral requirements (of which requirements the agent also is aware);\footnote{To
  capture the standard notion of a dilemma, the known courses of
  action must violate \textit{different} subsets of the agent's moral
  requirements.} and (2) is not immediately aware of a course of
action satisfying all moral requirements.\footnote{By `not aware' we roughly mean that
  the agent cannot immediately retrieve the information that some
  particular course of action will satisfy the moral requirements.
  Given a specific candidate solution, the agent may or may not be
  able to compute which moral requirements it satisfies; if so,
  a lack of `awareness' could result from having to search
  through too many possible plans before finding any correct solution
  that may exist.}
	
	
	We now describe a few factors which are significant to MQDs.
	
	\subsection{Solvability}
	
	MQDs may or may not be solvable, in that the
        agent may or may not actually be capable of some course of
        action which satisfies all moral requirements.  Importantly,
        for a situation to be considered a MQD, the agent
        must not know any solution to it. Further, the only ways to
        \textit{conclusively} determine if a MQD is solvable are to (a)
        find a solution, or (b) exhaustively search the space of all
        possible plans until all have been shown not to solve the
        problem, which will not be feasible in general.
	
	\subsection{Reason for uncertainty}
	
	In a MQD, all candidate solutions currently known by the agent violate some moral requirements.  If a MQD has a solution, this solution is unknown to the agent.  Solutions could be unknown either because the agent's action (or plan) space is so large that the agent cannot easily search all possible courses of action; or because the agent lacks information about the state of the world that affords a solution.\footnote{These reasons may be simultaneously active in a single MQD.}
	
	\subsection{Cognitive vs. physical exploration}
	
	In some MQDs the agent's search for solutions can be primarily cognitive (searching through the space of possible plans), with the search process having minimal impact on the agent's environment.  In other cases, the agent may need to \textit{physically} act on its environment in order to discover the means of solving the problem.\footnote{Whether cognitive/physical/both sorts of exploration are necessary is likely correlated with the reason for exploration --- MQDs due to partial state information are more likely to require physical exploration than those due to large plan spaces.}
	
	\subsection{Time pressure}
	
	Many moral dilemmas (such as most trolley problems) involve \textit{time pressure}: the agent  must choose a course of action within some time window, or else some unacceptable outcome will occur (five people will be hit by a trolley).  Time pressure remains an important factor in MQDs.
	
	When time pressure is a factor, it constrains the agent's ability to search (either cognitively or physically) for solutions.  Running out the clock trying to satisfy all moral requirements may be less permissible than selecting the lesser of known evils.  Any AMA that attempts to ``solve'' a MQD will likely need a mechanism to cut off such search with enough time to carry out the least immoral action seen so far.	
	
	\section{Example}\label{sec:example}
	
	We next introduce an additional example to illustrate how the aforementioned factors interact in a concrete MQD.
        
\begin{example2}{\it A military drone identifies a known terrorist, who will soon carry out a suicide attack that will kill twelve innocents.  The drone can target and kill the terrorist before the terrorist can carry out the attack, but its weapon's yield is too high to do so immediately without hitting four nearby civilians.  What should the drone do?}\end{example2}
        
	This scenario is one that an autonomous weapons system could conceivably face.  To some
        architectures, particularly those that treat targeting
        targeting decisions as fire/not fire, such a scenario would
        likely be treated as a moral dilemma  (see, e.g., Arkin's ethical governor\cite{arkin2009ethical}). It is in part the difficulty and starkness of such dilemmas that leads some to argue autonomous weapons should not be deciding between its two options at all.\footnote{Our inclusion of this MQD is not an endorsement of lethal autonomous weapons systems.} Regardless, the scenario should be regarded as a MQD.  The two most obvious courses of action (fire/do nothing) lead to morally unacceptable outcomes.  An agent may not know a course of action that would not result in civilian deaths.  Nevertheless, it is not inconceivable that some other course of action might satisfy all moral requirements (e.g., attempting to draw the terrorist away from the civilians); thus an agent that treats the scenario as a dilemma may entirely miss a morally preferable action.
	
	Whether this problem is solvable may not be clear even to outside observers. The uncertainty in this scenario likely arises both from a large plan space (a vast number of possible trajectories, so that not all can be considered) and hidden information about state (not knowing the terrorist's mental state means not knowing whether attempting to draw them away from civilians might succeed). Solving this MQD would likely require both cognitive and physical exploration: the agent may need consider non-obvious trajectories in order to investigate alternate angles of attack; evaluating whether attempts to draw the terrorist away would succeed may require actually attempting that action. Furthermore, time pressure matters: the extent to which the drone can search for a solution depends on how long it will be before the terrorist attacks.
	
	\section{Why use moral quasi-dilemmas in AMA development?}\label{sec:why}
	
	Humans are often faced with MQDs.  This is due to two features of the interaction between humans and their environment.
	
	First, humans have large plan spaces.  There are countless courses of action a human could perform even in one second: too large to possibly consider individually.  When faced with moral quandaries, humans may have immediate intuitions about which courses of action are  morally relevant, and may frame scenarios as moral dilemmas using these intuitions, but creative people may be able to transcend these circumscriptions and explore the broader plan space for solutions to moral quandaries.
	
	Second, humans necessarily have partial information about their environments. The human brain cannot store all information about anything that might become relevant.  Occasionally, some bit of unknown information about the world state may help resolve a moral quandary, such as when a hidden emergency brake could stop a speeding trolley from hitting people.
	
	Interactions between artificial agents (particularly robots, which operate in the physical world) and their environments will have similar characteristics.  Most robots have many degrees of freedom and can in principle generate a huge number of possible trajectories.  Additionally, robots will need to robustly interact with environments that are only partially observed.

	If the foregoing is true and artificial agents are likely to encounter MQDs ``in the wild'', then whether to handle these situations as moral dilemmas or to do something different is a significant question.
	To treat a MQD as a moral dilemma is to accept that the choice is between a limited number of actions, each of which violates some moral requirements.  Treating a solvable MQD as a dilemma guarantees that the agent will violate some moral requirements.  If some algorithm that attempts to explore the plan space or the physical world for a solution might find such a solution, then artificial agents that fail to do so when time is sufficient may be \textit{unnecessarily} violating moral requirements.
	If AMA designers ought to minimize the extent to which their creations violate moral requirements, then they ought to develop algorithms that consider MQDs and attempt to find solutions before concluding that doing so is impossible.
	
\section{How should AMAs handle moral quasi-dilemmas?}\label{sec:how}

When facing a MQD, how should an AMA respond?  In this section, we consider features AMAs may need in order to respond appropriately.

If the MQD is due to plan space intractability rather than partial information, then exploration is largely a cognitive endeavor.  Time pressure may constrain the agent so that there is some time at which the agent will need to stop searching and carry out the best plan found so far; continuing to explore at this stage would be much riskier.  However, the agent should likely explore the plan space for as long as possible subject to this constraint.  An agent that does not do so may allow a violation that could possibly have been avoided by finding a better action.

MQDs that are not resolvable without physical exploration are riskier.  Time constraints are again an issue, but the agent also runs the risk of performing an exploratory action that exacerbates the scenario (or, in a solvable MQD, renders the solution impracticable). The most acceptable course of action in such cases may indeed be to treat them as moral dilemmas, but some exploratory physical actions may still be obligatory, particularly when such actions are highly unlikely to hurt (e.g., the agent yelling and waving at railroad workers in the trolley problem).
	
In both cases, effective heuristics will be vital to effectively handling MQDs.  To maximize the probability of solving a MQD, the agent will need to effectively search the space of plans (and effectively estimate which exploratory actions are worth taking), focusing on the plans most likely to satisfy moral requirements.  Understanding human creative problem-solving, especially in moral domains, may help here.  Note that the agent could discover and subsequently pursue a course of action that itself might violate some moral requirements, provided the
violations are less severe than the originally available options.


Determining which action plans may be morally relevant may be considered an instance of the notoriously difficult frame problem.  This raises the question of whether effectively handling MQDs is too exacting a standard for evaluating artificial moral agents.  Though we should not expect AMAs to be able to solve all solvable MQDs within their respective time limits, we ought to design AMAs to \textit{attempt} to solve MQDs, as effectively as the state of the art allows.

\section{Conclusions and future work}\label{sec:conclusion}

In this paper we have defined and characterized the problem of MQDs, and argued for their utility in AMA development and evaluation.  We call for three lines of research in MQDs:

\begin{itemize}
	\item \textbf{Moral psychology and HRI research} to determine precisely how humans ascribe blame (both to other humans, and to robots/artificial agents) for exploration vs exploitation in MQDs.  Such research may also address how humans perceive MQDs when considering their own actions.
	\item \textbf{Formal definitions} both to characterize the notion of MQDs (e.g., in classical planning settings), and of specific MQDs.  This research will facilitate the use of MQDs for evaluating AMAs.  One possible approach might be to formalize MQDs as a subclass of what Sarathy and Scheutz call the ``MacGyver problem'', in which an agent must transcend its initial model of available actions and world states in order to achieve some goal.\cite{sarathy17arxiv}
	\item \textbf{Developing AMAs that handle MQDs}.  While probably no algorithm can solve every solvable MQD within its time constraints, we can at least develop architectures that support MQD handling.  We should then be able to incorporate continuing advances in computational creative problem solving and insights from cognitive science (such as bounded rationality\cite{Gigerenzer2010} and the explore-exploit tradeoff\cite{holland1975adaptation}) to improve such agents' capabilities.
\end{itemize}

\section*{Acknowledgements}

This project was supported in part by ONR MURI grant N00014-16-1-2278.

\bibliographystyle{ws-procs9x6}
\bibliography{icres2018-quasi}

\end{document}